%% file: neurips_2025.tex
\documentclass[final]{article}

\usepackage[dblblindworkshop]{neurips_2025}

\usepackage[utf8]{inputenc} 
\usepackage[T1]{fontenc}    
\usepackage{hyperref}       
\usepackage{url}            
\usepackage{booktabs}       
\usepackage{amsfonts}       
\usepackage{nicefrac}       
\usepackage{microtype}      
\usepackage{xcolor}         
\usepackage{algorithm}
\usepackage{algpseudocode}
\usepackage{amsmath}
\usepackage{float}
\usepackage{multirow}
\usepackage{siunitx}
\usepackage{array}
\usepackage[dvipsnames]{xcolor}
\usepackage{soul}
\usepackage{graphicx}

\newcommand{\giam}[1]{\textcolor{Maroon}{(#1)}}
\newcommand{\bt}[1]{\textcolor{YellowOrange}{(#1)}}
\newcommand{\tang}[1]{\textcolor{ForestGreen}{(#1)}}
\title{Metacognitive Sensitivity for Test-Time\\ Dynamic Model Selection}
\workshoptitle{CogInterp: Interpreting Cognition in Deep Learning Models}

\author{
    Le Tuan Minh Trinh \\
  University College London\\
     London, United Kingdom\\
  \texttt{minh.trinh.23@ucl.ac.uk} \\
  \And
  Le Minh Vu Pham \\
  FPT University \\
   Hanoi, Vietnam \\
  \texttt{vuplmhe180526@fpt.edu.vn} \\
  \AND
  Thi Minh Anh Pham \\
  Queen Mary University of London \\
  London, United Kingdom \\
  \texttt{t.pham@se22.qmul.ac.uk} \\
  \And
  Duc An Nguyen \\
  University of Oxford\\
  Oxford, United Kingdom \\
  \texttt{annguyen@robots.ox.ac.uk} \\
}

\begin{document}

\maketitle

\begin{abstract}
  A key aspect of human cognition is metacognition - the ability to assess one's own knowledge and judgment reliability. While deep learning models can express confidence in their predictions, they often suffer from poor calibration, a cognitive bias where expressed confidence does not reflect true competence. Do models truly know what they know? Drawing from human cognitive science, we propose a new framework for evaluating and leveraging AI metacognition. We introduce \textit{meta-d'}, a psychologically-grounded measure of metacognitive sensitivity, to characterise how reliably a model's confidence predicts its own accuracy. We then use this dynamic sensitivity score as context for a bandit-based arbiter that performs test-time model selection, learning which of several expert models to trust for a given task. Our experiments across multiple datasets and deep learning model combinations (including CNNs and VLMs) demonstrate that this metacognitive approach improves joint-inference accuracy over constituent models. This work provides a novel behavioural account of AI models, recasting ensemble selection as a problem of evaluating both short-term signals (confidence prediction scores) and medium-term traits (metacognitive sensitivity).

  
\end{abstract}

\input{sections/intro}
\input{sections/related-works}
\input{sections/methodology}

\input{sections/results}

\input{sections/discussions}

\bibliography{ref.bib}
\bibliographystyle{iclr2025_conference}
\appendix

\input{sections/appendix}
\end{document}

%% file: sections/intro.tex
\section{Introduction}

Deep learning research is characterised by increasing specialisation. Convolutional Neural Networks (CNNs) are widely used for perceptual tasks such as image recognition; Transformers and Large Language Models (LLMs) dominate natural language processing; and Vision-Language Models (VLMs) integrate across modalities~\cite{hu2024auxiliary,gadetsky2025large,mccoy2024embers, steyvers2025large}. This specialisation improves performance on benchmark tasks but also reflects the "No Free Lunch" theorem: no single architecture is optimal across all problems~\cite{bigelow2023context}. As a result, systems rely on multiple specialised models, raising the arbitration problem - determining which model should be used for a given input. This is the focus of dynamic model selection, which develops methods for assigning tasks to the most appropriate model.

A central obstacle in this area is the unreliability of model confidence. Deep neural networks produce probabilistic outputs that are often miscalibrated, meaning confidence does not align with true prediction accuracy. Research has shown that this miscalibration arises from the same architectural and training practices that enable high accuracy~\cite{guo2017calibration,geirhos2018imagenet,steyvers2025metacognition,song2025language}.

Cognitive science has long studied the analogous human ability, termed \textit{metacognition}, providing theoretical and mathematical tools for assessing how well an agent can evaluate its own knowledge~\cite{rouault2019forming,fleming2021know,lee2025metacognitive}. Concepts such as metacognitive sensitivity are increasingly viewed as necessary for developing AI systems that can self-monitor, detect errors, and adjust behaviour based on uncertainty~\cite{yedetore2023poor,kurvers2023automating,kuribayashi2025large,nguyen2025joint}. Recent work has applied the \textit{meta-d'} framework to AI, introducing "AI metacognitive sensitivity" as a performance dimension. Findings suggest that in human-AI collaboration, an AI with lower accuracy but higher metacognitive sensitivity can be a more effective partner, as its confidence signals are more reliable for guiding when its advice should be followed.


While the literature has started to recognise the importance of AI metacognition as a diagnostic tool~\cite{gandhi2024human,ivanova_elements_2024,gandhi2025cognitive,song_privileged_2025,murthy_inside_2025,song_language_2025,hu_re-evaluating_2025}, the work reviewed here represents a critical next step: operationalising this concept as a functional component within an adaptive algorithm for a concrete engineering problem like dynamic model selection at test-time. It moves the idea of AI metacognition from a desirable property to be measured to a dynamic signal to be actively leveraged.

In this work, we embed metacognitive sensitivity into a bandit-based framework for dynamic model selection, demonstrating accuracy gains compared to individual constituent models. Beyond these improvements, our results suggest that leveraging metacognition as a functional signal can enable ensemble systems that adapt more flexibly and reliably.


%% file: sections/related-works.tex
\section{Related Work}
Given a pool of expert models, a system must decide how to aggregate their outputs to arrive at a final prediction. The paradigms for combining and selecting models are summarised as follows:

\textbf{Static Ensembles:} combine predictions from all models using fixed rules, such as majority voting for classification or averaging for regression~\cite{ganaie2022ensemble}. 
Static ensembles reduce variance but are non-adaptive, treating all models as equally competent and using all of them for every input, which can be computationally inefficient when models have localised competence~\cite{ju2018relative,su2021adaptive,jiang2024interpretable}.

\textbf{Dynamic Ensembles Selection:} selects a model or subset of models for each test instance, based on the idea that different classifiers perform best in different regions of the feature space~\cite{cruz2015meta,liu2023dynamic,piwko2025divide,vasheghani2025dynamic}. 

\textbf{Mixture of Experts (MoE):} consists of expert networks and a gating network~\cite{chen2022towards}. The gating network assigns weights to experts for each input, and the final prediction is a weighted sum of expert outputs. Unlike Dynamic Selection, which uses pre-trained models with an external selection mechanism, MoE is trained end-to-end, with experts and gating network co-adapting during training~\cite{shazeer2017outrageously, li2024cumo,li2025uni,mu2025comprehensive}. 



The shift from static ensembles to dynamic selection, and MoE methods underscores the need for adaptive mechanisms. Existing methods rely on fixed heuristics like local accuracy rather than learned policies. Reinforcement learning, particularly contextual bandits, offers a framework for adaptive selection by balancing exploration and exploitation across models. However, current approaches lack metacognition-aware feedback. They either depend on miscalibrated raw confidence or on noisy, static heuristics such as local accuracy. None leverage measures of metacognitive sensitivity
to guide adaptive selection.

%% file: sections/methodology.tex
\section{Problem Formulation}

The central problem is to perform \textbf{test-time dynamic model selection} for an image classification task. Given a pair of pre-trained models, $M = \{M_A, M_B\}$, and a sequence of images, $D = \{x_1, \dots, x_N\}$, the objective is to create a framework that, for each image $x_t$, selects the model whose prediction is most likely to be correct.
Since models often vary in performance across domains, particularly on out-of-distribution data, a static choice is suboptimal. We propose a dynamic selection agent that chooses between models per sample, informed by both immediate confidence and a cognitively inspired, dynamically updated measure of \textbf{metacognitive sensitivity}~\cite{Fleming_Daw_2017}.

The goal is to learn a selection policy $\pi$ that maximises the cumulative reward, which is equivalent to maximising the total classification accuracy of the framework over the dataset: 
\begin{equation}
    \max_{\pi} \sum_{t=1}^{N} R_t = \max_{\pi} \sum_{t=1}^{N} \mathbb{I}(\hat{y}_{a_t, t} = y_t)
\end{equation}
where $a_t = \pi(s_t)$ is the model selected at time $t$ based on context $s_t$, $\hat{y}_{a_t, t}$ is its prediction, and $y_t$ is the ground truth label.
\section{Dynamic Model Selection Framework}
\vspace*{-5mm}
\begin{algorithm}[h]
\caption{Test-Time Dynamic Model Selection}
\label{alg:framework}
\begin{algorithmic}[1]
\State \textbf{Require:} Model set $M=\{M_A, M_B\}$, Dataset $D=\{x_1, \dots, x_N\}$, Burn-in size $B=100$, Window size $W=100$, Update frequency $F=50$.
\State \textbf{Initialise:} Contextual Bandit (LinUCB or LinTS).
\State \textbf{Initialise:} Total Reward $R_{total} \leftarrow 0$.

\Statex
\Comment{\textbf{Burn-in Phase}}
\State Collect performance data (confidence, reward) $H_k = \{(c_{k,i}, r_{k,i})\}_{i=1}^{B}$ for each model $M_k \in M$ on the first $B$ trials.
\For{each model $M_k \in M$}
    \State $\mu_{k,B} \leftarrow \text{meta-d'}(H_k)$ \Comment{Compute initial metacognitive score}
\EndFor

\Statex
\Comment{\textbf{Dynamic Selection Phase}}
\For{$t = B+1$ to $N$}
    \For{each model $M_k \in M$} \Comment{\textit{Context Formulation}}
        \State Get confidence $c_{k,t}$ on image $x_t$.
        \If{$(t-B) \pmod F = 1$}
            \State Collect past performance $H_k \leftarrow \{(c_{k,i}, r_{k,i})\}_{i=t-W}^{t-1}$.
            \State Update metacognitive score: $\mu_{k,t} \leftarrow \text{meta-d'}(H_k)$.
        \Else
            \State Keep previous score: $\mu_{k,t} \leftarrow \mu_{k,t-1}$.
        \EndIf
    \EndFor
    \State Construct context vector \begin{equation}
    s_t = [c_{A,t}, \mu_{A,t}, c_{B,t}, \mu_{B,t}]
    \end{equation} 
    
    \Statex
    \Comment{\textit{Bandit Action \& Reward}}
    \State Select model $a_t$ using bandit policy on $s_t$.
    \State $a_t \leftarrow \arg\max_{k \in \{A, B\}} \pi_t(s_t, k)$ \Comment{ $\pi_t(s_t, k)$: a specific bandit selection in Appendix}
    \State Get prediction $\hat{y}_{a_t, t}$ from model $M_{a_t}$.
    \State Observe true label $y_t$ and calculate reward $R_t = \mathbb{I}(\hat{y}_{a_t, t} = y_t)$.
    \State $R_{total} \leftarrow R_{total} + R_t$.
    
    \State Update bandit policy with $(s_t, a_t, R_t)$. \Comment{\textit{Learning}}
\EndFor

\State \Return Overall Accuracy: $R_{total} / (N-B)$.
\end{algorithmic}
\end{algorithm}

We propose a framework in which a contextual bandit serves as the dynamic selection agent. At each time step $t$, the bandit observes a context vector from two candidate models, selects one to act (the “arm”), receives a reward based on prediction correctness, and updates its policy. For each input $x_t$, models $M_A$ and $M_B$ produce predictions and confidence scores, which are combined into a 4-dimensional context vector as shown in Equation. 2.

The \textbf{short-term signal ($c_{k,t}$)} is the model's raw \textit{confidence} on the current image $x_t$, taken as the maximum value of its softmax output probability; and the \textbf{metacognitive sensitivity ($\mu_{k,t}$)} is the score representing the model's recent historical ability to know when it knows. This is a more stable, medium-term trait, which was operationalised using the \textbf{meta-d'} values. 

Sensitivity scores are initialised using the first 100 trials and updated every 50 trials via a 100-trial sliding window. To address the computational demands of hierarchical Bayesian inference, we developed a GPU-parallelised package for efficient \textit{meta-d'} estimation.
For learning and decision-making, we employ the contextual bandit algorithms 
(Further bandit algorithmic details are provided in the Appendix).

%% file: sections/results.tex
\section{Results}
We evaluate our framework using four diverse pre-trained models: AlexNet, GoogleNet (classical CNN), EfficientNet (efficiently scaled CNN), and Vision Transformer (ViT) (transformer-based model). This selection captures both convolutional and attention-based models, enabling the assessment of the adaptability of our framework across fundamentally different design choices. We evaluate accuracy at three checkpoints corresponding to
the number of trials indicates how many images have been processed up to that point.
\begin{table}[h]
    \caption{Accuracy of the best individual image models and the joint framework on the CIFAR10.}
    \centering
    \small
    \label{tab:image_model_performance}
    \begin{tabular}{l c c c c c c}
        \toprule
        Model Pair & \multicolumn{2}{c}{300 trials} & \multicolumn{2}{c}{700 trials} & \multicolumn{2}{c}{1000 trials} \\
        \cmidrule(lr){2-3} \cmidrule(lr){4-5} \cmidrule(lr){6-7}
         & {Model} & {Comb.} & {Model} & {Comb.} & {Model} & {Comb.} \\
        \midrule
        AlexNet-ViT  & 62.4 &69.5  \tang{+7.1\%} & 64.8 & 66.2 \bt{+1.4\%} & 62.4 & 65.9 \tang{+3.5\%} \\
         AlexNet-GoogleNet  & 62.7 & 70.6 \tang{+7.9\%} & 57.7 & 57.5 \giam{-0.2\%} & 56.8 & 58.4 \bt{+1.6\%} \\

       
        EfficientNet-ViT  & 67.7 & 75.9 \tang{+8.2\%} & 66.4 & 68.0 \bt{+1.6\%} & 66.4 & 67.8 \bt{+1.4\%} \\
       EfficientNet-GoogleNet  & 54.8 & 59.0 \tang{+4.8\%} & 53.6 & 55.8 \tang{+2.2\%} & 54.8 & 57.3 \tang{+2.5\%} \\
        \bottomrule
    \end{tabular}
\end{table}

Table~\ref{tab:image_model_performance} shows that the joint framework's accuracy is significantly improved in the early trials and then stabilises at 1.4\%–3.5\% higher than that of the individual models. At certain points, however, the joint framework underperforms relative to the best individual model in the pair. This effect can be attributed to correlated errors: performance is limited when both models misclassify the same sample, a pattern more common among architecturally similar models that share inductive biases from pre-training. For example, in one pairing (Figure.~\ref{fig:fig-alex-google}), the metacognitive sensitivity of AlexNet dropped (at trial 700), but the bandit quickly adapted by shifting selection toward GoogleNet. (Learning dynamics graphs are detailed in the Appendix~\ref{sec:appendix}.)

Despite these challenges, architectural diversity enhances complementarity. Heterogeneous pairings, such as CNN-Transformer models, exhibit fewer correlated errors and achieve higher accuracy. This highlights the importance of inductive-bias diversity for effective dynamic model selection.


\begin{table}[h]
    \centering
    \small
    \caption{Performance of the best individual VLM and the joint framework on the CIFAR10 - PACS.}
    \label{tab:vlms_performance}
    \begin{tabular}{l l c c c c c c}
        \toprule
        Model Pair & \multicolumn{2}{c}{1500 trials} & \multicolumn{2}{c}{2500 trials} & \multicolumn{2}{c}{4000 trials} \\
        \cmidrule(lr){2-3} \cmidrule(lr){4-5} \cmidrule(lr){6-7}
         & {Model} & {Comb.} & {Model} & {Comb.} & {Model} & {Comb.} \\
        \midrule
        \multirow{1}{*}{MetaCLIP-SigLIP} & 98.7 & 99.0 \bt{+0.3\%} & 98.7 & 98.6 \bt{0.0\%} & 98.4 & 98.5 \bt{+0.1\%} \\
        

        %
        \multirow{1}{*}{CLIP-ALIGN} & 94.2 & 96.0 \tang{+1.8\%} & 94.8 & 96.2 \bt{+1.6\%} & 94.8 & 95.8 \bt{+1.0\%} \\
        \bottomrule
    \end{tabular}
\end{table}

To further probe the framework’s robustness, we evaluated it under domain shift using Vision-Language Models (CLIP, ALIGN, SigLIP and MetaCLIP). To process the OOD bias in VLMs evaluation, we augment PACS~\cite{unknown} with CIFAR-10. PACS exhibits a 1:5 imbalance between photographic and non-photographic styles, which may skew the results. Adding CIFAR-10 introduces additional photographic diversity, yielding a more balanced and robust combined dataset for benchmarking.
Table~\ref{tab:vlms_performance} shows that while individual model performance declines under more demanding conditions, the framework leverages complementary strengths across models. This results in an initial accuracy improvement of 0.3\%–1.8\%, though because VLMs are already so accurate on their own, the gains are noticeable but modest compared to what had been seen with image models. However these potential suggest that metacognition-aware selection holds promise for addressing challenging, test-time distributional shifts in prediction tasks.




%% file: sections/discussions.tex
\section{Conclusion}

In this work, we presented a metacognition-inspired framework for dynamic model selection, introducing \textit{meta-d'} as a measure of how reliably a model’s confidence predicts its accuracy. 
Our experiments show that embedding metacognitive sensitivity in a bandit-based framework for dynamic model selection yields accuracy gains compared to individual constituent models. Future directions include extending the framework to large language model ensembles and exploring richer reinforcement learning strategies for selection. Overall, this work provides a first step toward incorporating metacognition-inspired feedback into adaptive model selection, with the goal of creating more reliable and interpretable ensemble systems.

%% file: sections/appendix.tex
\newpage
\section{Appendix}\label{sec:appendix}

\subsection{Metacognitive Sensitivity (\textit{meta-d'})}

To implement the metacognitive sensitivity score $\mu_k$, we adopt the `meta-d'` framework developed by \citeauthor{fleming2014measure} and detailed in \citet{Fleming_Daw_2017}. Metacognition refers to the ability to monitor one's own cognitive processes, and in this context, how well a model's confidence relates to its actual success.
\textit{Meta-d'} is a metric derived from \textbf{Signal Detection Theory (SDT)} by~\citeauthor{macmillan2002signal} that quantifies this ability. Its primary advantage is that it provides a measure of metacognitive sensitivity that is independent of the model's task performance and overall confidence bias.
We calculates \textit{meta-d'} by fitting a hierarchical Bayesian model to the observed distributions of confidence ratings for correct and incorrect trials~\cite{fleming2017hmeta}. For our framework, the score $\mu_{k,t}$ for a model $M_k$ at time $t$ is its calculated meta-d' value.

\subsection{Dynamic Update Mechanism}

A key feature of our framework is that the metacognitive sensitivity score, $\mu_k$, is not static but dynamically updated to reflect recent performance.
\begin{enumerate}
\item \textbf{Initialisation}: A burn-in set of the first 100 trials is used to compute the initial scores, $\mu_{A,0}$ and $\mu_{B,0}$.
\item \textbf{Sliding Window Update}: During the selection process, the \textit{meta-d'} score for each model is recalculated every 50 trials using a \textbf{sliding window} of the 100 most recent trials.
\end{enumerate}

This updating mechanism enables the framework to adapt to non-stationarity in model performance, such as when the underlying data distribution shifts.


\subsection{Contextual Bandit Selection}
The learning and decision-making core of our framework is a contextual bandit. We adopt two well-established algorithms: Algorithm.~\ref{alg:linucb}: \textbf{LinUCB} (Linear Upper Confidence Bound) and Algorithm.~\ref{alg:thompson}: \textbf{LinTS} (Linear Thompson Sampling). 
The bandit's interaction at each step $t$ is as follows:
\begin{enumerate}
    \item \textbf{Observe Context}: Receives the context vector $s_t$.
    \item \textbf{Select Action}: The algorithm's policy selects an action $a_t \in \{A, B\}$.
    \item \textbf{Receive Reward}: The framework executes the chosen model $M_{a_t}$. The reward $R_t$ is 1 if the model's prediction is correct, and 0 otherwise.
    \item \textbf{Update Policy}: The bandit uses the tuple $(s_t, a_t, R_t)$ to update its internal model, improving its policy for future decisions.
\end{enumerate}


\begin{algorithm}[H]
\caption{Linear Contextual Upper Confidence Bound}
\label{alg:linucb}
\begin{algorithmic}[1]
\Require Number of arms $K$, observe context $s_t$ with dimension $d$, exploration parameter $\alpha$
\State Initialise $A_a \gets I_d$ and $b_a \gets \mathbf{0}_d$ for all $k \in \{A,B\}$ 
\For{each round $t = 1, 2, \dots$}
    \State Observe context vector $s_t \in \mathbb{R}^d$
    \For{each arm $k \in \{A,B\}$}
        \State Compute $\hat{\theta}_k \gets A_k^{-1} b_k$
        \State Compute $\pi_t(s_t, k) \gets \hat{\theta}_k^\top s_t + \alpha \sqrt{s_t^\top A_k^{-1} s_t}$
    \EndFor
    \State Choose arm $k_t \leftarrow \arg\max_{k \in \{A, B\}} \pi_t(s_t, k)$
    
    \State Observe reward $r_t$
    \State Update:
        $A_{k_t} \gets A_{k_t} + s_t s_t^\top,\quad
        b_{k_t} \gets b_{k_t} + r_t s_t $

\EndFor
\end{algorithmic}
\end{algorithm}
\vspace*{-5mm}
\begin{algorithm}[H]
\caption{Linear Contextual Thompson Sampling}
\label{alg:thompson}
\begin{algorithmic}[1]
\Require Number of arms $K$, observe context $s_t$ with dimension $d$, prior variance parameter $\sigma$
\State Initialise $A_a \gets I_d$, $b_a \gets \mathbf{0}_d$ for all $k \in \{A,B\}$
\For{each round $t = 1, 2, \dots$}
    \State Observe context vector $s_t \in \mathbb{R}^d$
    \For{each arm $k \in \{A, B\}$}
        \State Compute $A_k^{-1} \gets (A_k + \epsilon I_d)^{-1}$ 
        \State Compute posterior mean: $\hat{\theta}_k \gets A_k^{-1} b_k$
        \State Sample $\tilde{\theta}_k \sim \mathcal{N}(\hat{\theta}_k, \sigma^2 A_k^{-1})$
        \State Compute sampled reward: $\pi_t(s_t, k) \gets \tilde{\theta}_k s_t^\top$
    \EndFor
    \State Choose arm $k_t \leftarrow \arg\max_{k \in \{A, B\}} \pi_t(s_t, k)$ $\rightarrow$ Observe reward $r_t$
    \State Update: $ A_{k_t} \gets A_{k_t} + s_t s_t^\top,\quad b_{k_t} \gets b_{k_t} + r_t s_t$
    
\EndFor
\end{algorithmic}
\end{algorithm}
\vspace*{-2mm}
\subsection{Learning Dynamics Graphs}

\begin{figure}[h]
    \centering
    \vspace*{-2mm}
    \includegraphics[width=1\linewidth]{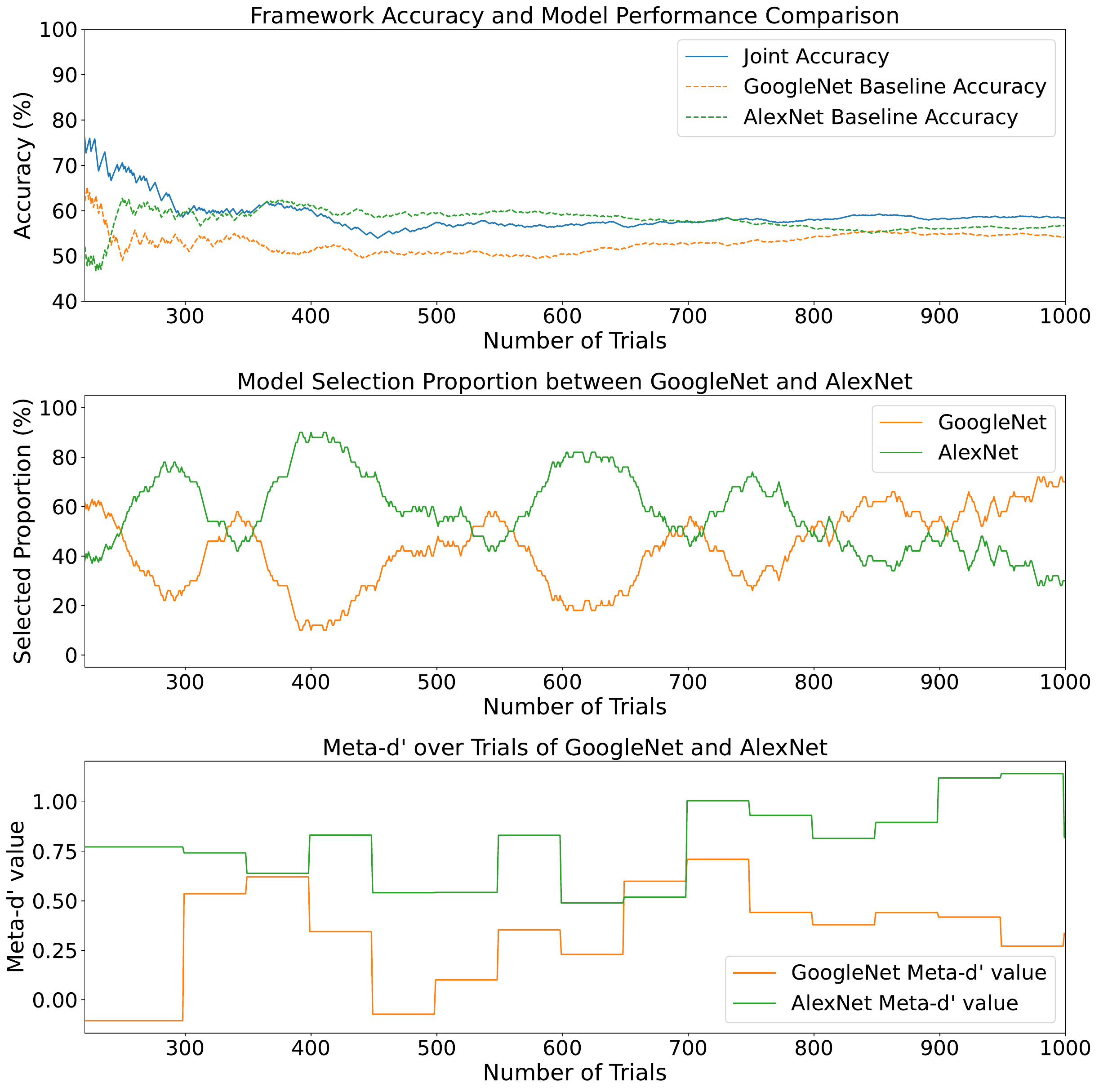}
    \caption{Figure of GoogleNet and AlexNet with Framework Accuracy (using LinTS with $\sigma$ = 0.5)}
    \label{fig:fig-alex-google}
\end{figure}

\begin{figure}
    \centering
    \includegraphics[width=1\linewidth]{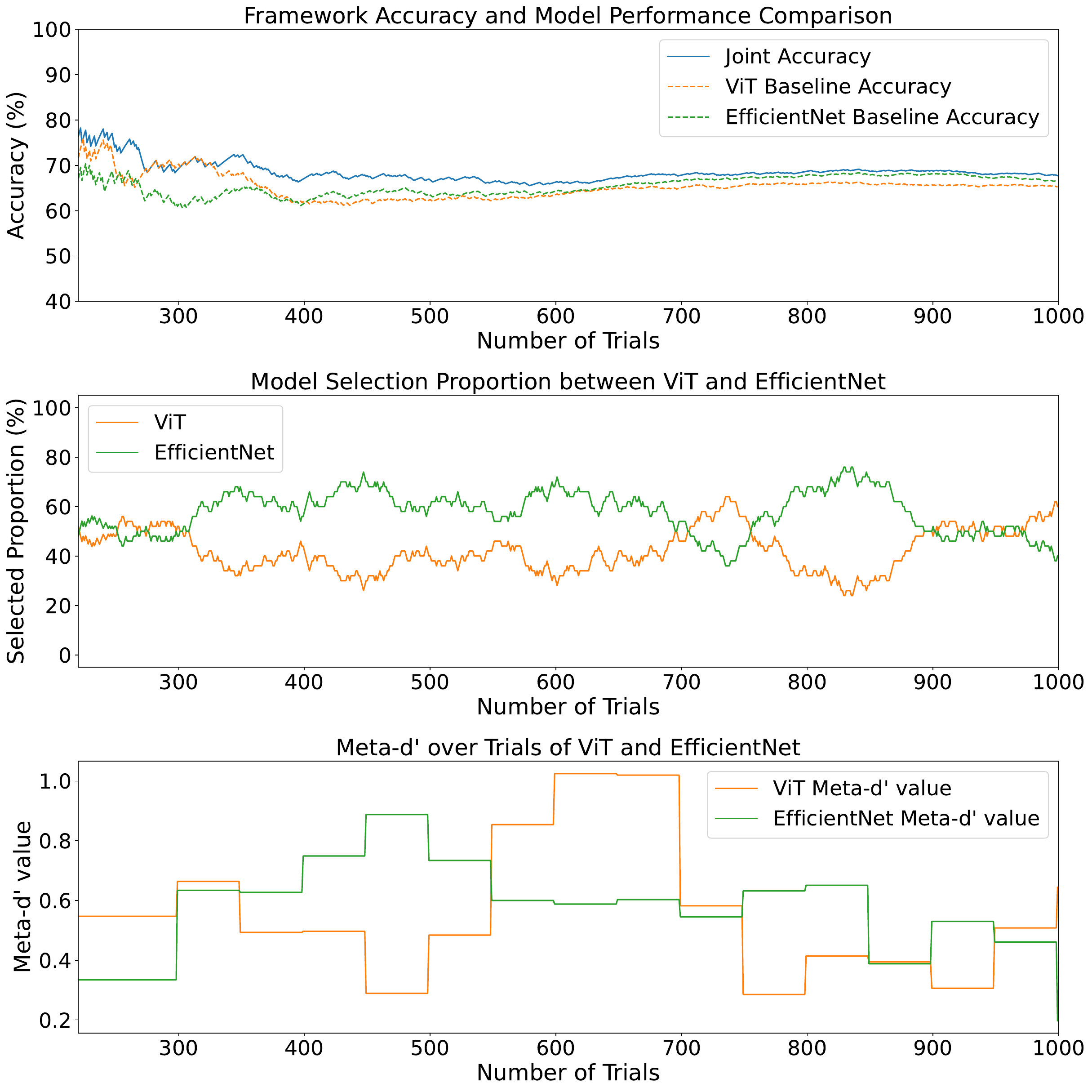}
    \caption{Figure of ViT and EfficientNet with Framework Accuracy (using LinTS with $\sigma$ = 1.0)}
    \label{fig:placeholder}
\end{figure}

\begin{figure}
    \centering
    \includegraphics[width=1\linewidth]{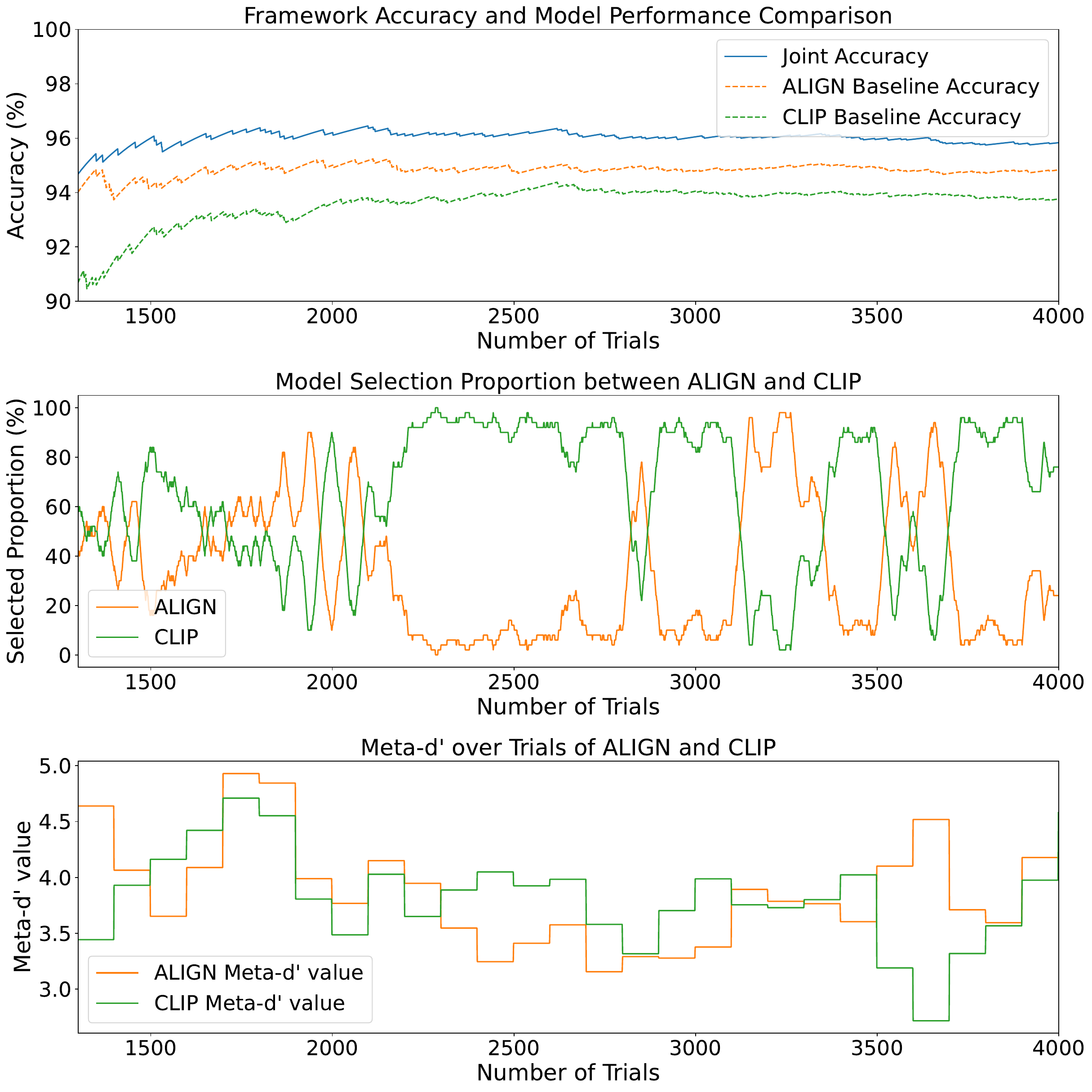}
    \caption{Figure of ALIGN and CLIP with Framework Accuracy (using LinUCB with $\alpha$ = 1.0)}
    \label{fig:placeholder}
\end{figure}

\begin{figure}
    \centering
    \includegraphics[width=1\linewidth]{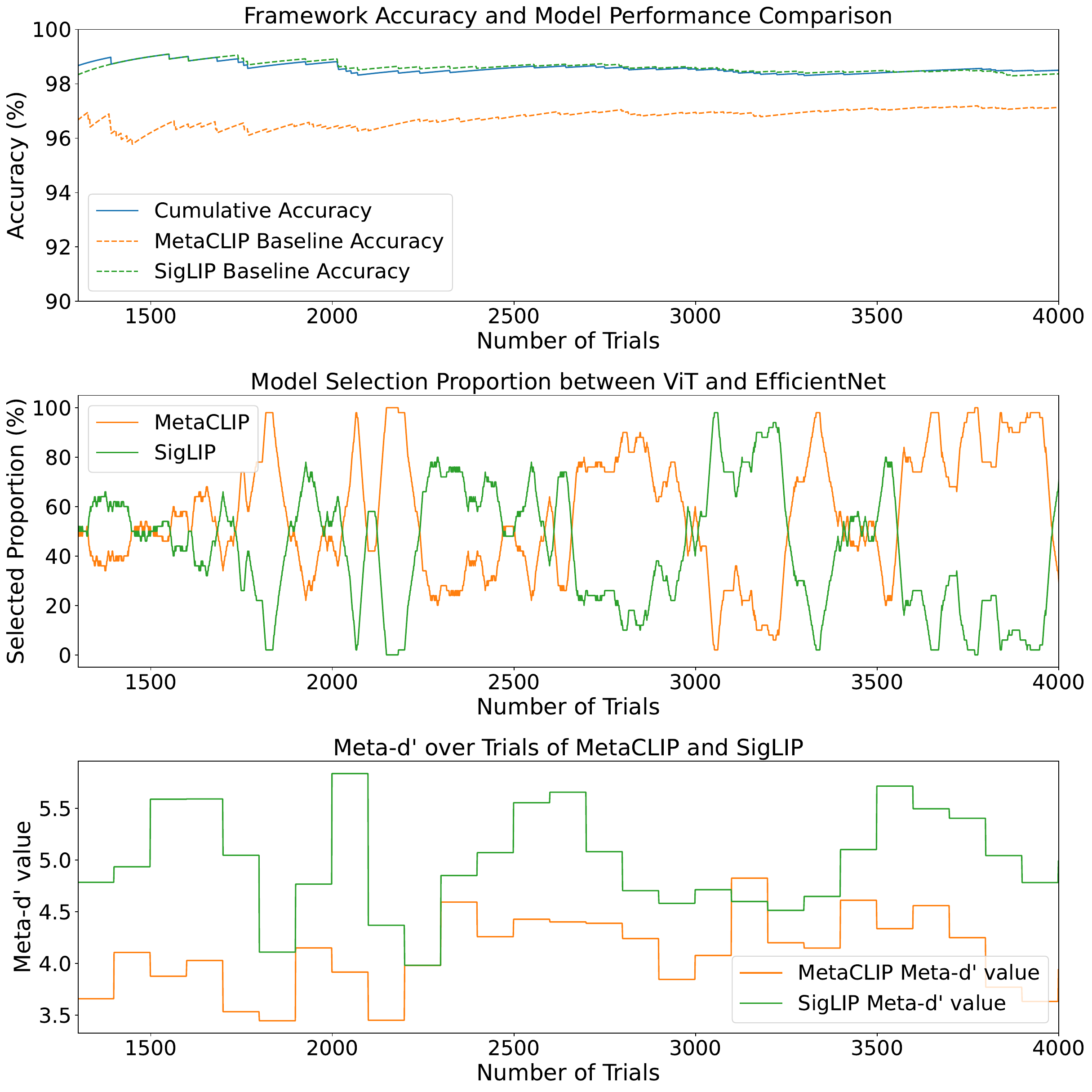}
    \caption{Figure of MetaCLIP and SigLIP with Framework Accuracy (using LinUCB with $\alpha$ = 0.5)}
    \label{fig:placeholder}
\end{figure}
\vspace*{-5mm}